\pdfoutput=1

\documentclass[runningheads]{llncs}
\usepackage{graphicx}
\usepackage{amsmath,amssymb} 
\usepackage{color,xcolor}
\usepackage[width=122mm,left=12mm,paperwidth=146mm,height=193mm,top=12mm,paperheight=217mm]{geometry}

\usepackage{booktabs}

\newcommand{\tbl}[1]{Table~\ref{#1}}

\DeclareRobustCommand\onedot{\futurelet\@let@token\@onedot}
\def\@onedot{\ifx\@let@token.\else.\null\fi\xspace}

\newcommand{\yasu}[1]{\textcolor{orange}{}}
\newcommand{\chen}[1]{\textcolor{blue}{}}
\newcommand{\charlie}[1]{\textcolor{green}{}}

\newcommand{\stress}[1]{\textcolor{orange}{{#1}}}
\newcommand{\stressnext}[1]{\textcolor{cyan}{{#1}}}

\begin{document}
\newcommand*\samethanks[1][\value{footnote}]{\footnotemark[#1]}

\pagestyle{headings}
\mainmatter
\def\ECCV18SubNumber{2518}  

\title{FloorNet: A Unified Framework for Floorplan Reconstruction from 3D Scans} 

\titlerunning{ECCV-18 submission ID \ECCV18SubNumber}

\authorrunning{ECCV-18 submission ID \ECCV18SubNumber}
\date{\vspace{-5ex}}
\author{
\makebox[.4\linewidth]{Chen Liu\thanks{The first two authors contribute equally on this work.} \qquad Jiaye Wu\samethanks} \\
Washington University in St. Louis \\
{\tt\small \{chenliu,jiaye.wu\}@wustl.edu}
\and
\makebox[.4\linewidth]{Yasutaka Furukawa} \\
Simon Fraser University \\
{\tt\small furukawa@sfu.ca}}


\maketitle



\begin{abstract}
The ultimate goal of this indoor mapping research is to automatically reconstruct a floorplan simply by walking through a house with a smartphone in a pocket. This paper tackles this problem by proposing FloorNet, a novel deep neural architecture. 
The challenge lies in the processing of RGBD streams spanning a large 3D space.
FloorNet effectively processes the data through three neural network branches: 1) PointNet with 3D points, exploiting the 3D information; 2) CNN with a 2D point density image in a top-down view, enhancing the local spatial reasoning; and 3) CNN with RGB images, utilizing the full image information. FloorNet exchanges intermediate features across the branches to exploit the best of all the architectures.
%
We have created a benchmark for floorplan reconstruction by acquiring RGBD video streams for 155 residential houses or apartments with Google Tango phones and annotating complete floorplan information.
Our qualitative and quantitative evaluations demonstrate that the fusion of three branches effectively improves the reconstruction quality. We hope that the paper together with the benchmark will be an important step towards solving a challenging vector-graphics reconstruction problem. Code and data are available at \url{https://github.com/art-programmer/FloorNet}.
\end{abstract}

\section{Introduction}
Architectural floorplans play a crucial role in designing, understanding, and remodeling indoor spaces. Their drawings are effective in conveying geometric and semantic information of a scene. For instance, we can quickly identify room extents, the locations of doors, or object arrangements (geometry). We can also recognize the types of rooms, doors, or objects easily through texts or icon styles (semantics).
Unfortunately, more than 90\% of houses in North America do not have floorplans. The ultimate goal of the indoor mapping research is to enable automatic reconstruction of a floorplan simply by walking through a house with a smartphone in a pocket.
%

The Consumer-grade depth sensors have revolutionized indoor 3D scanning  with successful products.
%
Matterport~\cite{matterport} produces detailed texture mapped models of indoor spaces by acquiring a set of panorama RGBD images with a specialized hardware. Google Project Tango phones~\cite{leegoogle} convert RGBD image streams into 3D or 2D models. These systems produce detailed geometry, but fall short as floorplans or architectural blue-prints, whose geometry must be concise and respect underlying scene segmentation and semantics.

Reconstruction of the floorplan for an entire house or an apartment with multiple rooms poses fundamental challenges to existing techniques due to its large 3D extent.
A standard approach projects 3D information onto a 2D lateral domain~\cite{ikehata2015structured}, losing the information of height. PointNet~\cite{qi2016pointnet,qi2017pointnet++}  consumes 3D information directly but suffers from the lack of local neighborhood structures.
A multi-view  representation~\cite{qi2016volumetric,su2015multi} avoids explicit 3D space modeling, but has been mostly demonstrated for objects, rather than large scenes and complex camera motions.
3D Convolutional Neural Networks (CNNs)~\cite{riegler2016octnet,tatarchenko2017octree} also show promising results but have been so far limited to objects or small-scale scenes.


This paper proposes a novel deep neural network (DNN) architecture FloorNet, which turns a RGBD video covering a large 3D space into pixel-wise predictions on floorplan geometry and semantics,
followed by an existing Integer Programming formulation~\cite{liu2017raster} to recover vector-graphics floorplans. FloorNet consists of three DNN branches. The first branch employs PointNet with 3D points, exploiting the 3D information. The second branch uses a CNN with a 2D point density image in a top-down floorplan view, enhancing the local spatial reasoning. The third branch uses a CNN with RGB images, utilizing the full image information. 
%
The PointNet branch and the point-density branch exchange features between the 3D points and their corresponding cells in the top-down view. The image branch contributes deep image features into the corresponding cells in the top-down view.
%
This hybrid DNN design exploits the best of all the architectures and effectively processes the full RGBD video covering a large 3D scene with complex camera motions.


We have created a benchmark for floorplan reconstruction by acquiring RGBD video streams for 155 residential houses or apartments with Google Tango phones and annotated their complete floorplan information including architectural structures, icons, and room types. 
Extensive qualitative and quantitative evaluates demonstrate the effectiveness of our approach over competing methods.
%

In summary, the main contributions of this paper are two-fold: 1) Novel hybrid DNN architecture for RGBD videos, which processes the 3D coordinates directly, models local spatial structures in the 2D domain, and incorporates the full image information; and 2) A new floorplan reconstruction benchmark with RGBD videos, where many indoor scene databases exist~\cite{dai2017scannet,song2016ssc,matterport} but none tackles a vector-graphics reconstruction problem, which has immediate impact on digital mapping, real estate, or civil engineering applications.


\section{Related work}
We discuss the related work in three domains: indoor scene reconstruction, 3D deep learning, and indoor scan datasets.

\vspace{0.2cm}
\noindent
\textbf{Indoor scene reconstruction:} 
The advancements in consumer-grade depth sensors have brought revolutionary changes to indoor 3D scanning.
KinectFusion~\cite{newcombe2011kinectfusion} enables high-fidelity 3D scanning for objects and small-scale scenes.
Whelan et al.~\cite{whelan2012kintinuous} extends the work to building-scale scans. While being accurate with details, these dense reconstructions fall short as CAD models, which must
have 1) concise geometry for efficient data transmission and 2) proper segmentations/semantics for architectural analysis or effective visualization.

Towards CAD-quality reconstructions,
researchers have applied model-based approaches by representing a scene with geometric primitives. Utilizing the 2.5D property of indoor building structures, rooms can be separated by fitting lines to points in a top-down view~\cite{okorn2010toward,turner2015fast,sui2016layer}. 
Primitive types have been extended to planes~\cite{furukawa2009manhattan,furukawa2009reconstructing,sinha2009piecewise,xiong2013automatic,mura2014automatic} or cuboids~\cite{xiao2014reconstructing}.
While they produce promising results for selected scans, they critically rely on the low-level geometry analysis for primitive detection, which faces challenges with noisy and incomplete 3D data. Our approach conducts global analysis of the entire input by DNNs to detect primitive structures much more robustly.

Another line of research studies the top-down scene reconstruction with shape grammars from a single image~\cite{zhao2011image} or a set of panorama RGBD images~\cite{ikehata2015structured,mura2016piecewise}.
Crowdsensing data such as images and WiFi-fingerprints are also exploited in building scene graphs~\cite{gao2014jigsaw,gao2016multi,luo2017constructing,jiang2013hallway}. 
While semantic segmentation~\cite{dai2017scannet,qi2016pointnet,qi2017pointnet++} and scene understanding~\cite{zhang2016deepcontext} are popular for indoor scenes, there has been no robust learning-based method for vector-graphics floorplan reconstruction. This paper provides such a method and its benchmark with the ground-truth.


One way to recover the mentioned vector-graphics floorplan models is from rasterized floorplan images~\cite{liu2017raster}. We share the same reconstruction target, and we utilize their Integer Programming formulation in our last step to recover the final floorplan. Nevertheless, instead of a single image as input, our input is a RGBD video covering a large 3D space, which requires a fundamentally different approach to  process the input data effectively.

\vspace{0.2cm}
\noindent
\textbf{3D deep learning:}
The success of CNN on 2D images has inspired research on 3D feature learning via DNNs.
Volumetric CNNs~\cite{wu20153d,maturana2015voxnet,qi2016volumetric} are  straightforward extensions of CNN to a 3D domain, but there are two main challenges: 1) data sparsity and  2) computational cost of 3D convolutions. FPNN~\cite{li2016fpnn} and Vote3D~\cite{wang2015voting} attempt to solve the first challenge, while OctNet~\cite{riegler2016octnet} and O-CNN~\cite{wang2017cnn} address the computational costs via octree representations.

2D CNNs with multi-view renderings have been successful for object recognition~\cite{qi2016volumetric,su2015multi} and
%
part segmentation~\cite{limberger2017shrec}. They effectively utilize all the image information but are so far limited to  regular (or fixed) camera arrangements. The extension to larger scenes with complex camera motions is not trivial.
%

PointNet~\cite{qi2016pointnet} directly uses 3D point coordinates to exploit the sparsity and avoid quantization errors, but it does not provide an explicit local spatial reasoning.
PointNet++~\cite{qi2017pointnet++} hierarchically groups points and adds spatial structures by 
farthest point sampling.
%
Kd-Networks~\cite{klokov2017escape} similarly group points by a KD-tree.
These techniques incur additional computational expenses due to the grouping and have been limited at object-scale.
For scenes, they need to split the space into smaller regions (e.g., 1m$\times$1m blocks) and process each region independently~\cite{qi2016pointnet,qi2017pointnet++}, potentially hurting global reasoning (e.g., identifying long walls for corridors or avoiding two kitchens for an apartment).


\vspace{0.2cm}
\noindent
\textbf{Indoor scan dataset:} 
Affordable depth sensing hardware enables researchers to build many indoor scan datasets.
The ETH3D dataset contains
only 16 indoor scans~\cite{schops2017multi}, and its purpose is for multi-view stereo rather than 3D point-cloud processing. The ScanNet dataset~\cite{dai2017scannet} and the SceneNN dataset~\cite{hua2016scenenn} capture a variety of indoor scenes. However, most of their scans contain only one or two rooms, not suitable for the floorplan reconstruction problem.

Matterport3D~\cite{chang2017matterport3d} builds high quality panorama RGBD image sets for 90 luxurious houses. 2D-3D-S dataset~\cite{armeni20163d} provides 6 large-scale indoor scans of office spaces by using the same Matterport camera. However, they focus on 2D and 3D semantic annotations, and do not address a vector-graphics reconstruction problem.
%
Meanwhile, they require an expensive specialized hardware (i.e., Matterport camera) for high-fidelity 3D scanning, while we aim to tackle the challenge by consumer-grade smartphones with low data quality.
%
%

Lastly, a large-scale synthetic dataset, SUNCG~\cite{song2016semantic}, offers a variety of indoor scenes with CAD-quality geometry and annotations. However, they are synthetic and cannot model the complexity of real scenes or replace the real photographs.
We provide the benchmark with full floorplan annotations and the corresponding RGBD videos from smartphones for 155 residential units.

\section{FloorNet}

The proposed FloorNet converts a RGBD video with camera poses into pixel-wise floorplan geometry and semantics information, which is an intermediate floorplan representation introduced by Liu et al.~\cite{liu2017raster}. We first explain the intermediate representation for being self-contained, then provide the details.

\subsection{Preliminaries}
The intermediate representation consists of the geometry and the semantics information.
The geometry part contains room-corners, object icon-corners, and door/window end-points, where the locations of each corner/point type are estimated by a 256$\times$256 heatmap in the 2D floorplan image domain, followed by a standard non-maximum suppression.
For example, a room corner is either I-, L-, T-, or X-shaped depending on the number of incident walls, making the total number of feature maps to be 13 considering their rotational variants. The semantics part is modeled as 1) 12 feature maps as a probability distribution function (PDF) over 12 room types, and 2) 8 feature maps as a PDF over 8 icon types.
We follow their approach and use their Integer Programming formulation to reconstruct a floorplan from this representation at the end.

%

\subsection{Triple-branch hybrid design} 
\begin{figure}[tb]
\centering
    \includegraphics[width=\linewidth]{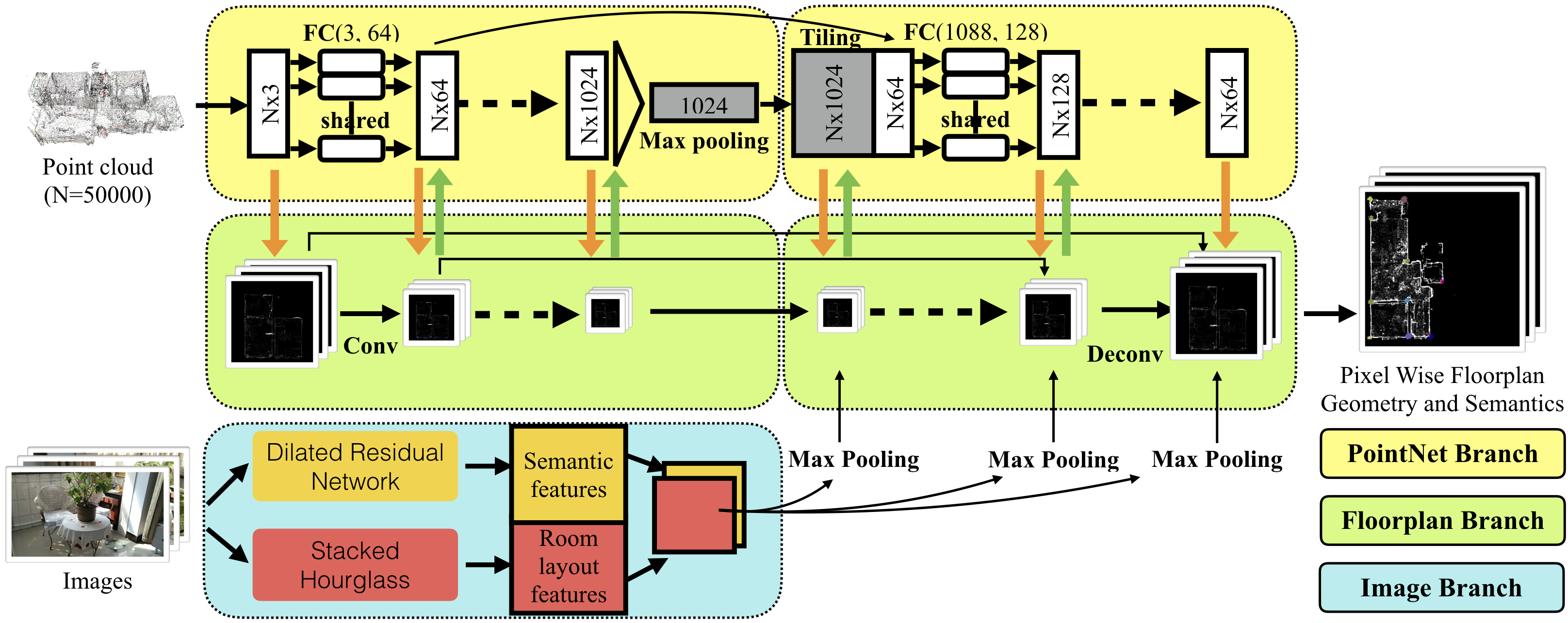}
\caption{FloorNet consists of three DNN branches. The first branch uses PointNet~\cite{qi2016pointnet} to directly consume 3D information. The second branch takes a top-down point density image in a floorplan domain with a fully convolutional network~\cite{long2015fully}, and produces pixel-wise geometry and semantics information. The third branch produces deep image features by a dilated residual network trained on the semantic segmentation task~\cite{yu2017dilated} as well as a stacked hourglass CNN trained on the room layout estimation~\cite{newell2016stacked}. The PointNet branch and the floorplan branch exchanges intermediate features at every layer, while the image branch contributes deep image features into the decoding part of the floorplan branch. This hybrid DNN architecture effectively processes an input RGBD video with camera poses, covering a large 3D space.
%
}
    \label{fig:pipeline}
    \vspace{-5pt}
\end{figure}

Floornet consists of three DNN branches. We employ existing DNN architectures in each branch without modifications. Our contribution lies in its hybrid design: how to combine them and share intermediate features (See Fig.~\ref{fig:pipeline}).

\vspace{0.2cm}
\noindent
{\bf PointNet Branch:} The first branch is PointNet~\cite{qi2016pointnet} that directly takes 3D points, where we use the original architecture without modifications.
We randomly subsample 50,000 points for each data. We 
manually rectify the rotation before annotation, while aligning the gravity direction with the Z-axis. We add translation to move the center of mass to the origin. 

\vspace{0.2cm}
\noindent
{\bf Floorplan Branch:} The second branch is a fully convolutional network (FCN)~\cite{long2015fully} with skip connections between the encoder and the decoder, which takes a point-density image in the top-down view. We compute a 2D axis-aligned bounding box of the Manhattan-rectified 3D points to define a rectangular floorplan domain, while ignoring the 2.5\% outlier points and expanding the rectangle by 5\% in each of the four directions. The rectangle is placed in the middle of the 256$\times$256 square image in which the geometry and semantics feature maps are produced. The input to the branch is a point-density image in the same domain.


\vspace{0.2cm}
\noindent
{\bf Image Branch:} The third branch computes deep image features through two CNN architectures: 1) Dilated residual network (DRN)~\cite{yu2017dilated} trained on semantic segmentation; and 2) stacked hourglass CNN (HG)~\cite{newell2016stacked} trained on room layout estimation. We have used ScanNet~\cite{dai2017scannet} benchmark to train DRN and LSUN~\cite{zhang2015large} benchmark to train HG.



\subsection{Intra-branch feature sharing}
\begin{figure}[tb]
\centering
    \includegraphics[width=0.85\linewidth]{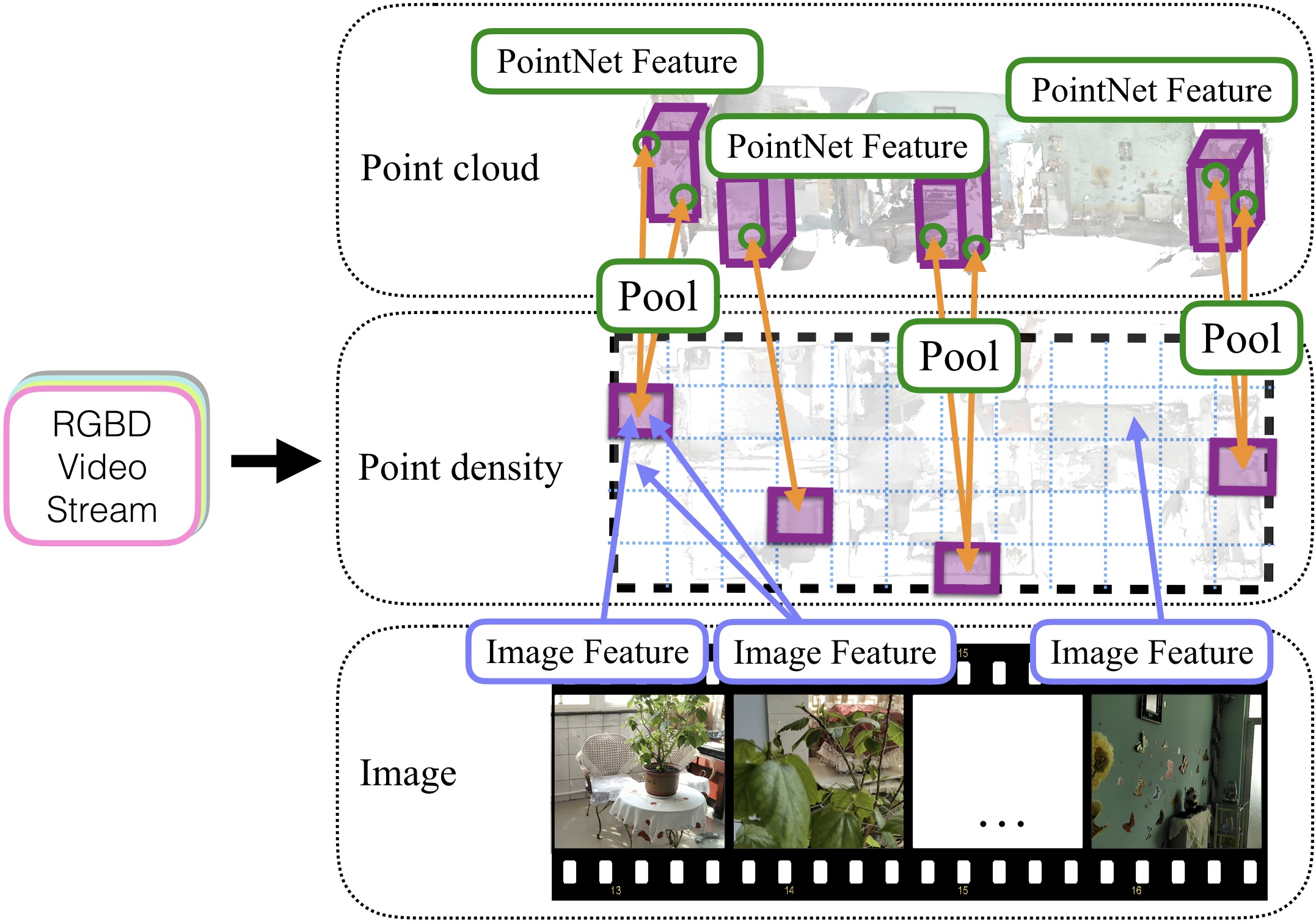}
\caption{FloorNet shares features across branches to exploit the best of all the architectures. PointNet features at 3D points are pooled into corresponding 2D cells in the floorplan branch. Floorplan features at 2D cells are unpooled to the corresponding 3D points in the PointNet branch. Deep image features are pooled into corresponding 2D cells in the floorplan branch, based on the depthmap and the camera pose information.
%
}
    \label{fig:fusion}
    \vspace{-5pt}
\end{figure}

Different branches learn features in different domains (3D points, the floorplan, and images). FloorNet offers three intra-branch feature sharing by pooling and unpooling operations, based on the camera poses and 3D information (See Fig.~\ref{fig:fusion}).
%

\vspace{0.2cm}
\noindent
{\bf PointNet to floorplan pooling:} This pooling module takes features of unordered points
from each layer of the PointNet branch and produces a 2D top-down feature map in the corresponding layer of the floorplan branch. The module simply spreads point features into cells defined by the output top-down feature map, then computes the sum of the features in each cell.
%
%
Though the projection could be performed along each of the three axes, 
we focus on the vertical projection since our goal is to reconstruct a floorplan.
%
%
%
%
%
A constructed feature map has the same dimension as the layer in the floorplan branch, and is simply concatenated to the current feature stack.
%
%
%
The time complexity of the projection pooling module is linear in the number of 3D points.

\vspace{0.2cm}
\noindent
{\bf Floorplan to pointNet unpooling:} 
This module reverses the above pooling operation. It 
simply copies and adds a feature of the floorplan cell into each of the corresponding 3D points that project inside the cell.
%
The time complexity is again linear in the number of points. 

\vspace{0.2cm}
\noindent
{\bf Image to floorplan pooling:} 
The image branch produces two deep image features of dimensions 512x32x32 and 256x64x64 from DRN and HG for each video frame. We first unproject image features to 3D space by their depthmaps and camera poses, and then apply the same 3D to floorplan pooling operation to aggregate 3D features to the floorplan branch. We conduct the image branch pooling for every 10 frames in the video sequence.

\subsection{Loss functions}
Our network outputs pixel-wise predictions on the floorplan geometry and semantics information in the same resolution
$256\times 256$. 
For geometry heatmaps (i.e., room corners, object icon-corners, and door/window end-points), a sigmoid cross entropy loss is used. The ground-truth heatmap is prepared by putting a value of 1.0 inside a disk of radius 11 pixels around each ground-truth pixel.
For semantic classification feature maps (i.e., room types and object icon types), a pixel-wise softmax cross entropy loss is used.

\section{Floorplan reconstruction benchmark} \label{section:benchmark}

\begin{figure}[tb]
\includegraphics[width=\linewidth]{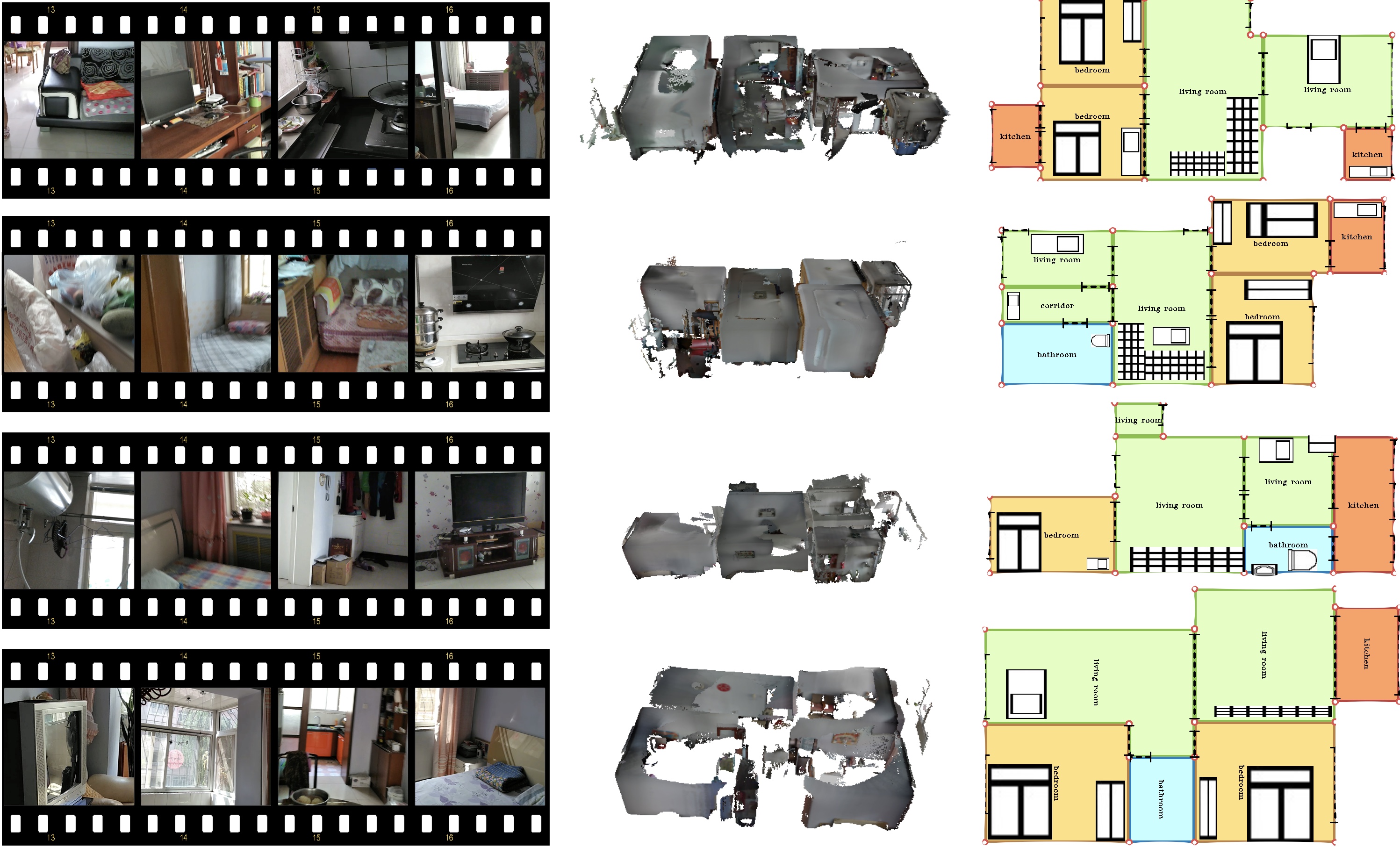}
\caption{Floorplan reconstruction benchmark. From left to right: Subsampled video frames, colored 3D point clouds, and ground-truth  floorplan data. The floorplan data is stored in a vector-graphics representation, which is visualized with a simple rendering engine (e.g., rooms are assigned different colors based on their types, and objects are shown as canonical icons).
}\label{fig:dataset}
\end{figure}

This paper creates a benchmark for the vector-graphics floorplan reconstruction problem from RGBD videos with camera poses.
We have acquired roughly two-hundreds 3D scans of residential units in the United States and China using Google Tango phones (Lenovo Phab 2 Pro and Asus ZenFone AR)
(See Fig.~\ref{fig:dataset}). After manually removing poor quality scans, we have annotated the complete floorplan information for the remaining 155 scans: 1) room-corners as points, 2) walls as pairs of room-corners, 3) object icons and types as axis-aligned rectangles and classification labels, 4) doors and windows (i.e., openings) as line-segments on walls, and 5) room types as classification labels for polygonal areas enclosed by walls. The list of object types is \{\textit{counter, bathtub, toilet, sink, sofa, cabinet, bed, table, refrigerator}\}. The list of room types is \{\textit{living room, kitchen, bedroom, bathroom, closet, balcony, corridor, dining room}\}. \tbl{tbl:statistics} provides statistics of our data collections.
\begin{table}[t]
\centering
\caption{Dataset statistics. From left to right: the number of rooms, the number of icons, the number of openings (i.e., doors or windows), the number of room-corners, and the total area. The average and the standard deviation are reported for each entry.}
\label{tbl:statistics}
\begin{tabular}{c|c|c|c|c|c}
\toprule
  & \#room & \#icon &\#opening & \#corner & area \\
\midrule
Average & 5.2 & 9.1 &  9.9 & 18.1 & 63.8[$m^2$] \\
\hline
Std & 1.8 & 4.5 & 2.9 & 4.2 & 13.0[$m^2$] \\
\bottomrule
\end{tabular}
\end{table}


Reconstructed floorplans are evaluated on three different levels of geometric and semantic consistency with the ground-truth. 
We follow the work by Liu et al.~\cite{liu2017raster} and define the low- and mid-level metrics as follows.

\noindent $\bullet$ The low-level metric is the precision and recall of room corner detections. A corner detection is declared a success if its distance to the ground-truth is below 10 pixels and the closest among all the other room corners.

\noindent $\bullet$ The mid-level metric is the precision and recall of detected openings (i.e., doors and windows), object-icons, and rooms. The detection of an opening is declared a success if the largest distance of the corresponding end-points is less than 10 pixels. The detection of an object (resp. a room) is declared a success if the intersection-over-union (IOU) with the ground-truth is above 0.5 (resp. 0.7).

\noindent $\bullet$
Relationships of architectural components play crucial roles in evaluating indoor spaces. For example, one may look for apartments where bedrooms are not connected to a kitchen. A building code may enforce every bedroom to have a quick evacuation route to outside through windows or doors in case of fire. We introduce the high-level metric as the ratio of rooms that have the correct relationships with the neighboring rooms.
More precisely, we declare that a room has a correct relationship if 1) a room is connected to the correct set of rooms through doors, where two rooms are connected if their common walls contain at least one door, 2) each room (i.e., the room and its neighboring rooms) has an IOU score larger than 0.5 with the corresponding ground-truth, and 3) each room has the correct room type.



\section{Implementation details}
\subsection{DNN Training}
Among the 155 scans we collected, we randomly sample 135 for training and leave 20 for testing. We perform data augmentation by random scaling and rotation 
every time we feed a training sample. First, we apply rescaling to the point-cloud and the corresponding annotation with a random factor uniformly sampled from a range $[0.5, 1.5]$. Second,  
we randomly apply the rotation around the z axis by either $0^o$, $90^o$, $180^o$, or $270^o$. 

We use the official code for each DNN module, that is, 
PointNet~\cite{qi2016pointnet},
FCN~\cite{long2015fully} for the Floorplan branch, and DRN~\cite{yu2017dilated} and SH~\cite{newell2016stacked} for the Image branch.
We pre-train DRN on the semantic segmentation task with ScanNet database~\cite{dai2017scannet} and SH on the room layout estimation task with LSUN~\cite{zhang2015large}.
DRN and SH are fixed during the FloorNet training, and we optimize only the PointNet and the floorplan branches. 

FloorNet has three types of loss functions. We have observed that enabling all the loss functions in a single training would lead to worse testing performance, because the model overfits with the icon loss.
Instead, we train the model with each loss function one by one. When using the icon loss, we limit the training to be at most 600 iterations and use early-stopping based on the testing loss, where 1 iteration contains 20 batches.~\footnote{ 
We incorporated synthetic dataset SUNCG~\cite{song2016semantic} and/or real dataset Matterport3D~\cite{chang2017matterport3d} for training with the icon loss, while using their semantic segmentation information to produce icon annotations. However, the joint-training still experiences overfitting, while this simply early-stopping heuristic works well in our experiments.}
%

Training of FloorNet takes around 2 hours with a TitanX GPU. On the average, the training continues for about 600 iterations, consuming $1,620,000= 135 (\mbox{samples})\times 600 (\mbox{iterations}) \times 20 (\mbox{batches})$ augmented training samples. It is initially to our surprise that FloorNet generalizes even from a small number of 3D scans. However, FloorNet makes pixel-wise predictions, which are mostly low-level vision tasks. Each 3D scan contains about 10 object-icons, 10 openings, and a few dozen room corners, which probably lead to the good generalization performance together with data augmentation, where similar phenomena were observed by Chen et al.~\cite{liu2017raster}

\subsection{Enhancement heuristics} 

We augment the Integer Programming Formulation~\cite{liu2017raster} with the following two enhancement heuristics to deal with more challenging input data (i.e., large-scale raw sensor data) and hence more noise in the network predictions.

\vspace{0.2cm}
\noindent {\bf Primitive candidate generation:}
Standard non-maximum suppression often detects multiple room corners around a single ground-truth. 
After thresholding the room-corner heatmap by a value 0.5, we simply extract the highest peak from each connected component, whose area is more than 5 pixels.
%
To handle localization errors, we connect two room-corners and generate a wall candidate when their corresponding connected components overlap along X or Y direction.
We do not augment junctions to keep the number of candidates tractable.

\vspace{0.2cm}
\noindent {\bf Objective function:}
Wall and opening candidates are originally assigned uniform weights in the objective function~\cite{liu2017raster}. We calculate the confidence of a wall (resp. opening) candidate by taking the average of the semantic heatmap scores of type "wall" along the line with width 7 pixels (resp. 5 pixels).
We set the weight of each primitive by the confidence score minus 0.5, so that a primitive
is encouraged to be chosen only when the confidence is at least 0.5.

\section{Experiments}

\begin{figure}[tb]
\includegraphics[width=\linewidth]{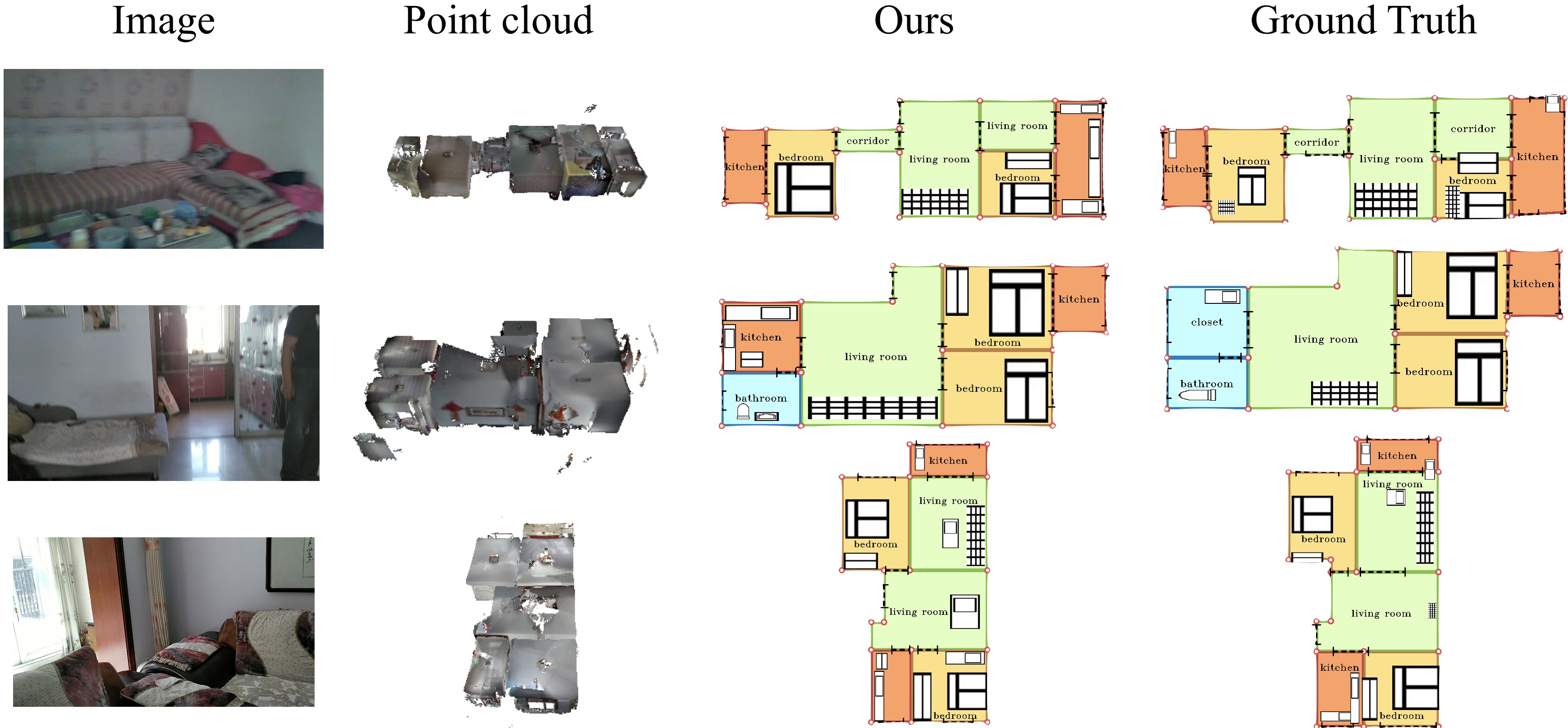}
\caption{Floorplan reconstruction results.
}
\label{fig:results}
\end{figure}

\begin{figure}[p]
\includegraphics[width=\linewidth]{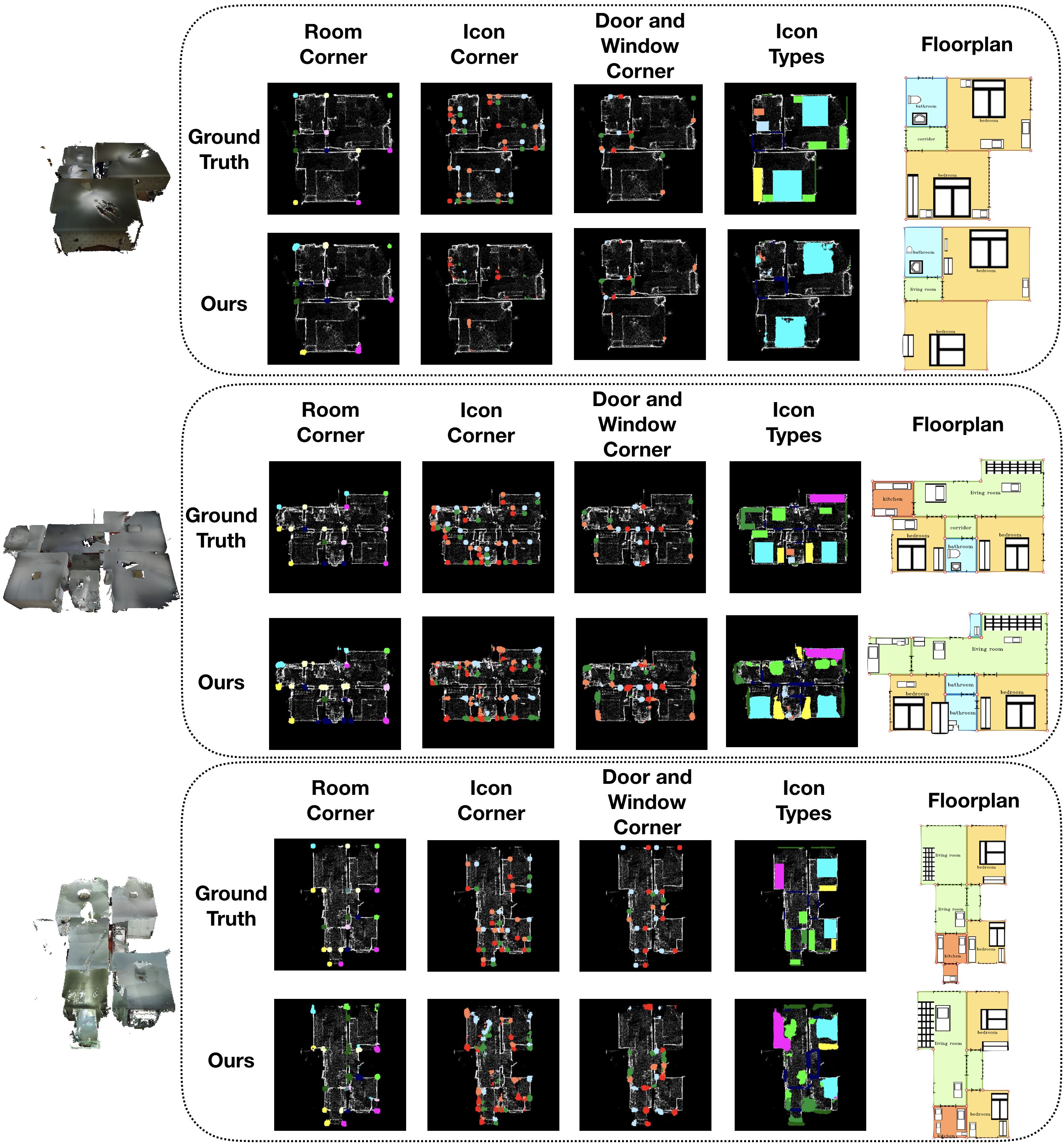}
\caption{Intermediate results. For each example, we show raw outputs of the networks (room corners, icon corners, opening corners, and icon types) compared against the ground-truth.
%
%
In the second example, we produce a fake room (blue color) at the top due to poor quality 3D points. In the third example, reconstructed rooms have inaccurate shapes near the bottom left again due to noisy 3D points, illustrating the challenge of our problem.
}
\label{fig:intermediate}
\end{figure}

\begin{figure}[p]
\centerline{
\includegraphics[width=\linewidth]{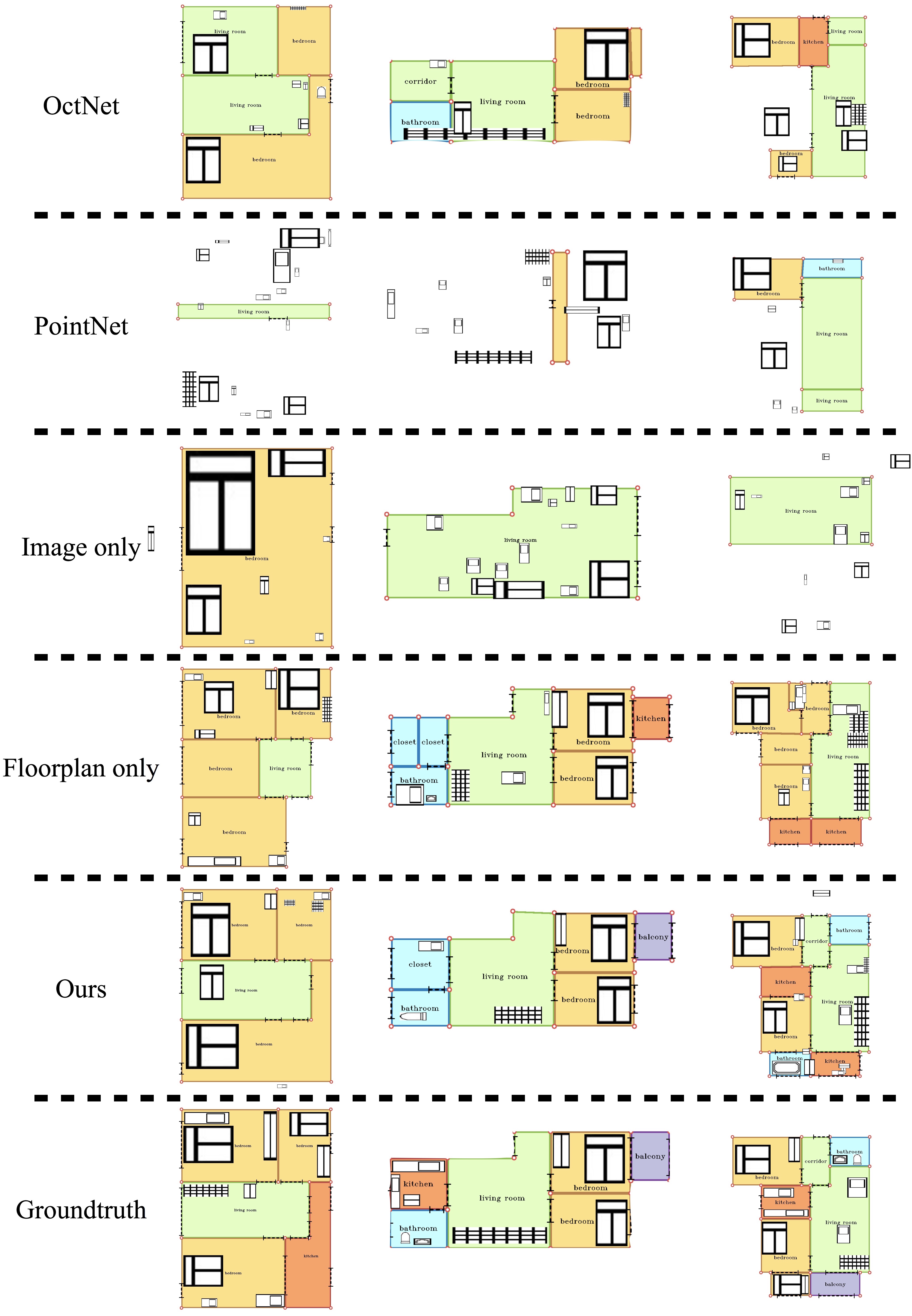}
}
\vspace{-0.5cm}
\caption{Qualitative comparisons against competing methods. The top is OctNet~\cite{riegler2016octnet}, a state-of-the-art 3D CNN architecture. The next three rows show variants of our FloorNet, where only one branch is enabled. FloorNet with all the branches overall produce more complete and accurate floorplans.
}\label{fig:compare}
\end{figure}

\begin{table}[tb]
\caption{Quantitative evaluations on low-, mid-, and high-level metrics against competing methods and our variants.
The orange and cyan color indicates the best and the second best result for each entry.
}
\label{tbl:results}
\begin{center}
\begin{tabular}{l|c||c|c|c||c}
\hline
 & wall & door & icon & room & relationship\\\hline\hline
PointNet~\cite{qi2016pointnet} & 25.8/42.5 & 11.5/38.7 & 22.5/27.9 & 27.0/40.2 & 5.0\\
\hline
Floorplan-branch & 90.2/88.7 & 70.5/78.0 & 43.4/42.8 & 76.3/75.3 & 50.0\\
\hline
Image-branch & 40.0/83.3 & 15.4/47.1 & 21.4/17.4 & 25.0/57.1 & 0.0\\
\hline
OctNet~\cite{riegler2016octnet} & 75.4/89.2 & 36.6/\stressnext{82.3} & 32.8/48.8 & 62.1/72.0 & 13.5\\
\hline
\hline
Ours w/o PointNet-Unpooling & \stress{92.6}/92.1 & 75.8/76.8 & \stressnext{55.1}/51.9 & 80.9/77.4 & 52.3\\
\hline
Ours w/o PointNet-Pooling & 88.4/\stress{93.0} & 73.0/\stress{87.2} & 50.0/42.2 & 75.0/80.6 & 52.8\\
\hline
Ours w/o Image-Pooling & \stress{92.6}/89.7 & \stress{77.1}/74.4 & 50.5/\stress{57.8} & \stress{84.2}/\stressnext{83.1} & \stress{56.8}\\
\hline
Ours & 92.1/\stressnext{92.8} & \stressnext{76.7}/80.2 & \stress{56.1}/\stress{57.8} & \stressnext{83.6}/\stress{85.2} & \stress{56.8} \\
\hline
\end{tabular}
\end{center}
\end{table}

Figure~\ref{fig:results} shows our reconstruction results on some representative examples. Our approach successfully recovers complex vector-graphics floorplan data including room geometries and their connectivities through doors. One of the major failure modes is in the icon detection. As the model training experiences overfitting in the icon loss, object detection generally requires more training data than low-level geometry (i.e., corners) detection~\cite{liu2017raster}. We believe that more training data will overcome this issue. Another typical failures come from missing room corners due to clutter or incomplete scanning. The successful reconstruction of a room requires successful detection of every single room corner. This is a challenging problem and the introduction of higher level constraints may reveal a solution (e.g., if one sees a door, a room must be reconstructed on the other side even with missing corners).


Figure~\ref{fig:compare} and Table~\ref{tbl:results} qualitatively and quantitatively compare our method against competing techniques, namely, OctNet~\cite{riegler2016octnet}, PointNet~\cite{qi2016pointnet}, and a few variants of our FloorNet.
OctNet and PointNet represent state-of-the-art 3D DNN techniques.
More precisely, we implement the voxel semantic segmentation network based on the official OctNet library,~\footnote{OctNet library: https://github.com/griegler/octnet} which takes 256x256x256 voxels as input and outputs 3D voxels of the same resolution. We then add three separate $5\times 3 \times 3$ convolution layers with strides $4\times1\times1$ to predict the same pixel-wise geometry and semantics feature-maps with
the same set of loss functions. 
PointNet is simply our FloorNet without the point density or the image input.
%
Similarly, we construct a FloorNet variant by enabling only the 3D points (for the PointNet branch) or the point density image (for the floorplan branch) as the input.

The table shows that the floorplan branch is the most informative as it is the most natural representation for floorplan reconstruction task, while PointNet branch or Image branch alone does not work well.
We also split the entire point clouds into $1m\times 1m$ blocks, train the PointNet-only model that makes predictions per block separately, followed by a simple merging. However, this performs much worse.
OctNet performs reasonably well across low- to mid-level metrics, but does poorly on the high-level metric, where all the rooms and relevant doors must be reconstructed at high precision to report good numbers.


To further evaluate the effectiveness of the proposed FloorNet architecture, we conduct ablation studies by disabling each of the intra-branch pooling/unpooling operations. The bottom of Table~\ref{tbl:results} shows that the feature sharing overall leads to better results, especially for mid- to high-level metrics.

\begin{figure}
\centerline{
\rotatebox{90}{\hspace{0.5cm}Tango}
\includegraphics[width=0.9\linewidth]{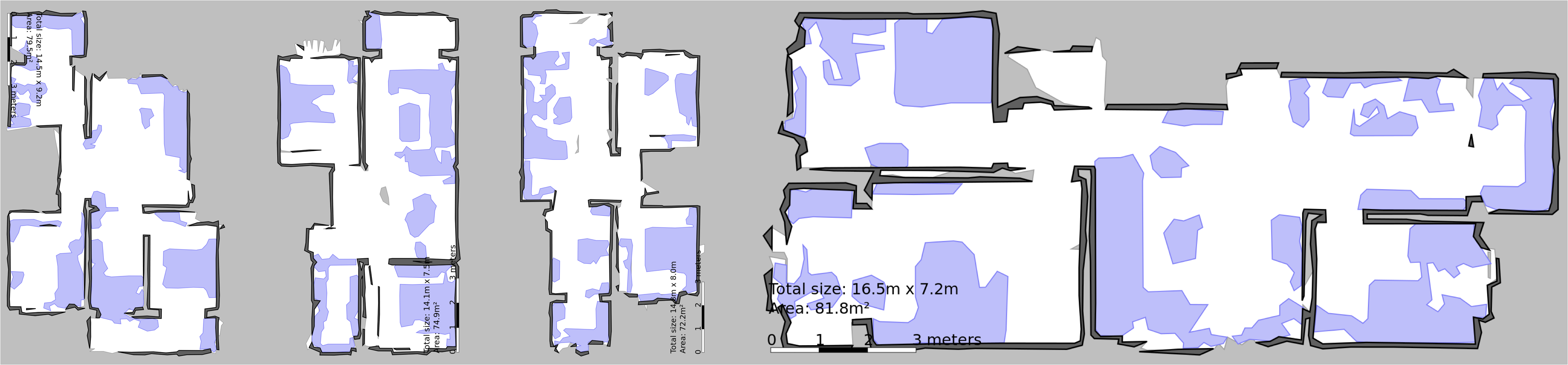}
}
\vspace{0.1cm}
\centerline{
\rotatebox{90}{\hspace{0.7cm}Ours}
\includegraphics[width=0.9\linewidth]{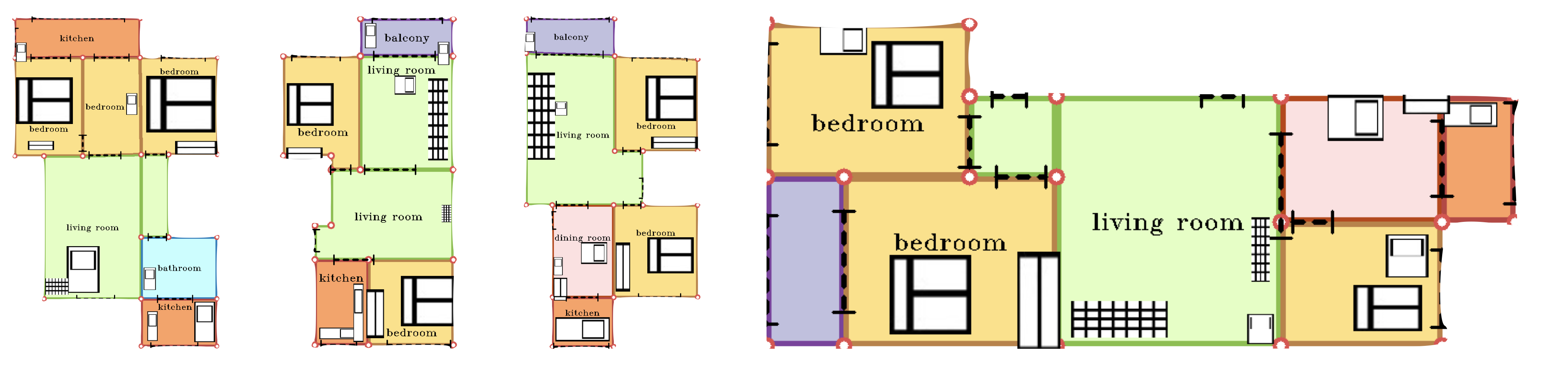}
}
\caption{Comparison against a commercial floorplan generator, Tango Navigator App. Top: Floorplan image from Tango. Bottom: Our results.
}
\label{fig:tango}
\end{figure}

Finally, Figure~\ref{fig:tango} compares against a build-in Tango Navigator App~\cite{tango}, which generates a floorplan image real-time on the phone. Note that their system does not 1) produce room segmentations, 2) recognize room types, 3) detect objects, 4) recognize object types, or 5) produce CAD-quality geometry. Therefore,
we quantitatively evaluate only the geometry information by 
measuring the line distances between the ground-truth walls and predicted walls. More precisely, 
we 1) sample 100 points from each wall line segment, 2) for each sampled point, find the closest one in the other line segment, and 3) compute the mean distance over all the sampled points and line segments. The average line distances are 2.72 [pixels] and 1.66 [pixels] for Tango Navigator App and our FloorNet, respectively. This is a surprisingly result to some extent, because our algorithm drops many confident line segments during Integer Programming, when their corresponding rooms miss one corner and are not reconstructed.
On the other hand, it is an expected result as our approach makes full use of geometry and image information to recover a floorplan.




\section{Conclusion}
This paper proposes a novel DNN architecture FloorNet that reconstructs vector-graphics floorplans from RGBD videos with camera poses. FloorNet takes a hybrid approach and exploits the best of three DNN architectures to effectively process a RGBD video covering a large 3D space with complex camera motions.
The paper also provides a new benchmark for a new vector-graphics reconstruction problem, which is missing in the recent indoor scene databases of Computer Vision.
Two main future works are ahead of us. The first one is to learn to enforce higher level constraints inside DNNs as opposed to inside a separate post-processing (e.g., Integer Programming). Learning high-level constraints likely require more training data and the second future work is to acquire more scans.

More than 90\% of houses in North America do not have floorplans. We hope that this paper together with the benchmark will be an important step towards solving this challenging vector-graphics reconstruction problem, and enabling the reconstruction of a floorplan just by walking through a house with a smartphone. We will publicly share our code and data to promote further research.

\section*{Acknowledgement}
This research is partially supported by National Science Foundation under grant IIS 1540012 and IIS 1618685, Google Faculty Research Award, and Adobe gift fund. We thank Nvidia for a generous GPU donation.

\clearpage

\bibliographystyle{splncs}
\bibliography{main}

\begin{thebibliography}{10}

\bibitem{matterport}
:
\newblock Matterport.
\newblock https://matterport.com/

\bibitem{leegoogle}
Lee, J., Dugan, R.,  et~al.:
\newblock Google project tango

\bibitem{ikehata2015structured}
Ikehata, S., Yang, H., Furukawa, Y.:
\newblock Structured indoor modeling.
\newblock In: Proceedings of the IEEE International Conference on Computer
  Vision. (2015)  1323--1331

\bibitem{qi2016pointnet}
Qi, C.R., Su, H., Mo, K., Guibas, L.J.:
\newblock Pointnet: Deep learning on point sets for 3d classification and
  segmentation.
\newblock arXiv preprint arXiv:1612.00593 (2016)

\bibitem{qi2017pointnet++}
Qi, C.R., Yi, L., Su, H., Guibas, L.J.:
\newblock Pointnet++: Deep hierarchical feature learning on point sets in a
  metric space.
\newblock In: Advances in Neural Information Processing Systems. (2017)
  5105--5114

\bibitem{qi2016volumetric}
Qi, C.R., Su, H., Nie{\ss}ner, M., Dai, A., Yan, M., Guibas, L.J.:
\newblock Volumetric and multi-view cnns for object classification on 3d data.
\newblock In: Proceedings of the IEEE Conference on Computer Vision and Pattern
  Recognition. (2016)  5648--5656

\bibitem{su2015multi}
Su, H., Maji, S., Kalogerakis, E., Learned-Miller, E.:
\newblock Multi-view convolutional neural networks for 3d shape recognition.
\newblock In: Proceedings of the IEEE international conference on computer
  vision. (2015)  945--953

\bibitem{riegler2016octnet}
Riegler, G., Ulusoys, A.O., Geiger, A.:
\newblock Octnet: Learning deep 3d representations at high resolutions.
\newblock arXiv preprint arXiv:1611.05009 (2016)

\bibitem{tatarchenko2017octree}
Tatarchenko, M., Dosovitskiy, A., Brox, T.:
\newblock Octree generating networks: Efficient convolutional architectures for
  high-resolution 3d outputs.
\newblock arXiv preprint arXiv:1703.09438 (2017)

\bibitem{liu2017raster}
Liu, C., Wu, J., Kohli, P., Furukawa, Y.:
\newblock Raster-to-vector: Revisiting floorplan transformation.
\newblock In: Proceedings of the IEEE Conference on Computer Vision and Pattern
  Recognition. (2017)  2195--2203

\bibitem{dai2017scannet}
Dai, A., Chang, A.X., Savva, M., Halber, M., Funkhouser, T., Nie{\ss}ner, M.:
\newblock Scannet: Richly-annotated 3d reconstructions of indoor scenes.
\newblock In: Proc. IEEE Conf. on Computer Vision and Pattern Recognition
  (CVPR). Volume~1. (2017)

\bibitem{song2016ssc}
Song, S., Yu, F., Zeng, A., Chang, A.X., Savva, M., Funkhouser, T.:
\newblock Semantic scene completion from a single depth image.
\newblock IEEE Conference on Computer Vision and Pattern Recognition (2017)

\bibitem{newcombe2011kinectfusion}
Newcombe, R.A., Izadi, S., Hilliges, O., Molyneaux, D., Kim, D., Davison, A.J.,
  Kohi, P., Shotton, J., Hodges, S., Fitzgibbon, A.:
\newblock Kinectfusion: Real-time dense surface mapping and tracking.
\newblock In: Mixed and augmented reality (ISMAR), 2011 10th IEEE international
  symposium on, IEEE (2011)  127--136

\bibitem{whelan2012kintinuous}
Whelan, T., Kaess, M., Fallon, M., Johannsson, H., Leonard, J., McDonald, J.:
\newblock Kintinuous: Spatially extended kinectfusion.
\newblock (2012)

\bibitem{okorn2010toward}
Okorn, B., Xiong, X., Akinci, B., Huber, D.:
\newblock Toward automated modeling of floor plans.
\newblock In: Proceedings of the symposium on 3D data processing, visualization
  and transmission. Volume~2. (2010)

\bibitem{turner2015fast}
Turner, E., Cheng, P., Zakhor, A.:
\newblock Fast, automated, scalable generation of textured 3d models of indoor
  environments.
\newblock IEEE Journal of Selected Topics in Signal Processing \textbf{9}(3)
  (2015)  409--421

\bibitem{sui2016layer}
Sui, W., Wang, L., Fan, B., Xiao, H., Wu, H., Pan, C.:
\newblock Layer-wise floorplan extraction for automatic urban building
  reconstruction.
\newblock IEEE transactions on visualization and computer graphics
  \textbf{22}(3) (2016)  1261--1277

\bibitem{furukawa2009manhattan}
Furukawa, Y., Curless, B., Seitz, S.M., Szeliski, R.:
\newblock Manhattan-world stereo.
\newblock In: Computer Vision and Pattern Recognition, 2009. CVPR 2009. IEEE
  Conference on, IEEE (2009)  1422--1429

\bibitem{furukawa2009reconstructing}
Furukawa, Y., Curless, B., Seitz, S.M., Szeliski, R.:
\newblock Reconstructing building interiors from images.
\newblock In: Computer Vision, 2009 IEEE 12th International Conference on, IEEE
  (2009)  80--87

\bibitem{sinha2009piecewise}
Sinha, S., Steedly, D., Szeliski, R.:
\newblock Piecewise planar stereo for image-based rendering.
\newblock (2009)

\bibitem{xiong2013automatic}
Xiong, X., Adan, A., Akinci, B., Huber, D.:
\newblock Automatic creation of semantically rich 3d building models from laser
  scanner data.
\newblock Automation in Construction \textbf{31} (2013)  325--337

\bibitem{mura2014automatic}
Mura, C., Mattausch, O., Villanueva, A.J., Gobbetti, E., Pajarola, R.:
\newblock Automatic room detection and reconstruction in cluttered indoor
  environments with complex room layouts.
\newblock Computers \& Graphics \textbf{44} (2014)  20--32

\bibitem{xiao2014reconstructing}
Xiao, J., Furukawa, Y.:
\newblock Reconstructing the world’s museums.
\newblock International journal of computer vision \textbf{110}(3) (2014)
  243--258

\bibitem{zhao2011image}
Zhao, Y., Zhu, S.C.:
\newblock Image parsing with stochastic scene grammar.
\newblock In: Advances in Neural Information Processing Systems. (2011)  73--81

\bibitem{mura2016piecewise}
Mura, C., Mattausch, O., Pajarola, R.:
\newblock Piecewise-planar reconstruction of multi-room interiors with
  arbitrary wall arrangements.
\newblock In: Computer Graphics Forum. Volume~35., Wiley Online Library (2016)
  179--188

\bibitem{gao2014jigsaw}
Gao, R., Zhao, M., Ye, T., Ye, F., Wang, Y., Bian, K., Wang, T., Li, X.:
\newblock Jigsaw: Indoor floor plan reconstruction via mobile crowdsensing.
\newblock In: Proceedings of the 20th annual international conference on Mobile
  computing and networking, ACM (2014)  249--260

\bibitem{gao2016multi}
Gao, R., Zhao, M., Ye, T., Ye, F., Luo, G., Wang, Y., Bian, K., Wang, T., Li,
  X.:
\newblock Multi-story indoor floor plan reconstruction via mobile crowdsensing.
\newblock IEEE Transactions on Mobile Computing \textbf{15}(6) (2016)
  1427--1442

\bibitem{luo2017constructing}
Luo, H., Zhao, F., Jiang, M., Ma, H., Zhang, Y.:
\newblock Constructing an indoor floor plan using crowdsourcing based on
  magnetic fingerprinting.
\newblock Sensors \textbf{17}(11) (2017)  2678

\bibitem{jiang2013hallway}
Jiang, Y., Xiang, Y., Pan, X., Li, K., Lv, Q., Dick, R.P., Shang, L., Hannigan,
  M.:
\newblock Hallway based automatic indoor floorplan construction using room
  fingerprints.
\newblock In: Proceedings of the 2013 ACM international joint conference on
  Pervasive and ubiquitous computing, ACM (2013)  315--324

\bibitem{zhang2016deepcontext}
Zhang, Y., Bai, M., Kohli, P., Izadi, S., Xiao, J.:
\newblock Deepcontext: Context-encoding neural pathways for 3d holistic scene
  understanding.
\newblock arXiv preprint arXiv:1603.04922 (2016)

\bibitem{wu20153d}
Wu, Z., Song, S., Khosla, A., Yu, F., Zhang, L., Tang, X., Xiao, J.:
\newblock 3d shapenets: A deep representation for volumetric shapes.
\newblock In: Proceedings of the IEEE Conference on Computer Vision and Pattern
  Recognition. (2015)  1912--1920

\bibitem{maturana2015voxnet}
Maturana, D., Scherer, S.:
\newblock Voxnet: A 3d convolutional neural network for real-time object
  recognition.
\newblock In: Intelligent Robots and Systems (IROS), 2015 IEEE/RSJ
  International Conference on, IEEE (2015)  922--928

\bibitem{li2016fpnn}
Li, Y., Pirk, S., Su, H., Qi, C.R., Guibas, L.J.:
\newblock Fpnn: Field probing neural networks for 3d data.
\newblock In: Advances in Neural Information Processing Systems. (2016)
  307--315

\bibitem{wang2015voting}
Wang, D.Z., Posner, I.:
\newblock Voting for voting in online point cloud object detection.
\newblock In: Robotics: Science and Systems. (2015)

\bibitem{wang2017cnn}
Wang, P.S., Liu, Y., Guo, Y.X., Sun, C.Y., Tong, X.:
\newblock O-cnn: Octree-based convolutional neural networks for 3d shape
  analysis.
\newblock ACM Transactions on Graphics (TOG) \textbf{36}(4) (2017) ~72

\bibitem{limberger2017shrec}
Limberger, F.A., Wilson, R.C., Aono, M., Audebert, N., Boulch, A., Bustos, B.,
  Giachetti, A., Godil, A., Le~Saux, B., Li, B.,  et~al.:
\newblock Shrec'17 track: point-cloud shape retrieval of non-rigid toys.
\newblock In: 10th Eurographics workshop on 3D Object retrieval. (2017)  1--11

\bibitem{klokov2017escape}
Klokov, R., Lempitsky, V.:
\newblock Escape from cells: Deep kd-networks for the recognition of 3d point
  cloud models.
\newblock In: 2017 IEEE International Conference on Computer Vision (ICCV),
  IEEE (2017)  863--872

\bibitem{schops2017multi}
Sch{\"o}ps, T., Sch{\"o}nberger, J.L., Galliani, S., Sattler, T., Schindler,
  K., Pollefeys, M., Geiger, A.:
\newblock A multi-view stereo benchmark with high-resolution images and
  multi-camera videos.
\newblock In: Proc. CVPR. Volume~3. (2017)

\bibitem{hua2016scenenn}
Hua, B.S., Pham, Q.H., Nguyen, D.T., Tran, M.K., Yu, L.F., Yeung, S.K.:
\newblock Scenenn: A scene meshes dataset with annotations.
\newblock In: 3D Vision (3DV), 2016 Fourth International Conference on, IEEE
  (2016)  92--101

\bibitem{chang2017matterport3d}
Chang, A., Dai, A., Funkhouser, T., Halber, M., Nie{\ss}ner, M., Savva, M.,
  Song, S., Zeng, A., Zhang, Y.:
\newblock Matterport3d: Learning from rgb-d data in indoor environments.
\newblock arXiv preprint arXiv:1709.06158 (2017)

\bibitem{armeni20163d}
Armeni, I., Sener, O., Zamir, A.R., Jiang, H., Brilakis, I., Fischer, M.,
  Savarese, S.:
\newblock 3d semantic parsing of large-scale indoor spaces.
\newblock In: Proceedings of the IEEE Conference on Computer Vision and Pattern
  Recognition. (2016)  1534--1543

\bibitem{song2016semantic}
Song, S., Yu, F., Zeng, A., Chang, A.X., Savva, M., Funkhouser, T.:
\newblock Semantic scene completion from a single depth image.
\newblock arXiv preprint arXiv:1611.08974 (2016)

\bibitem{long2015fully}
Long, J., Shelhamer, E., Darrell, T.:
\newblock Fully convolutional networks for semantic segmentation.
\newblock In: Proceedings of the IEEE conference on computer vision and pattern
  recognition. (2015)  3431--3440

\bibitem{yu2017dilated}
Yu, F., Koltun, V., Funkhouser, T.:
\newblock Dilated residual networks.
\newblock In: Computer Vision and Pattern Recognition. Volume~1. (2017)

\bibitem{newell2016stacked}
Newell, A., Yang, K., Deng, J.:
\newblock Stacked hourglass networks for human pose estimation.
\newblock In: European Conference on Computer Vision, Springer (2016)  483--499

\bibitem{zhang2015large}
Zhang, Y., Yu, F., Song, S., Xu, P., Seff, A., Xiao, J.:
\newblock Large-scale scene understanding challenge: Room layout estimation.
\newblock accessed on Sep \textbf{15} (2015)

\bibitem{tango}
Inc., G.:
\newblock Project tango.
\newblock https://developers.google.com/tango/

\end{thebibliography}
\end{document}